# Deep Latent Dirichlet Allocation with Topic-Layer-Adaptive Stochastic Gradient Riemannian MCMC

Yulai Cong [1]  Bo Chen [1]  Hongwei Liu [1]  Mingyuan Zhou [2]

## Abstract

It is challenging to develop stochastic gradient based scalable inference for deep discrete latent variable models (LVMs), due to the difficulties in not only computing the gradients, but also adapting the step sizes to different latent factors and hidden layers. For the Poisson gamma belief network (PGBN), a recently proposed deep discrete LVM, we derive an alternative representation that is referred to as deep latent Dirichlet allocation (DLDA). Exploiting data augmentation and marginalization techniques, we derive a block-diagonal Fisher information matrix and its inverse for the simplex-constrained global model parameters of DLDA. Exploiting that Fisher information matrix with stochastic gradient MCMC, we present topic-layer-adaptive stochastic gradient Riemannian (TLASGR) MCMC that jointly learns simplex-constrained global parameters across all layers and topics, with topic and layer specific learning rates. State-of-the-art results are demonstrated on big data sets.

## 1. Introduction

The increasing amount and complexity of data call for large-capacity models, such as deep discrete latent variable models (LVMs) for unsupervised data analysis (Hinton et al., 2006; Bengio et al., 2007; Srivastava et al., 2013; Ranganath et al., 2015; Zhou et al., 2016a), and scalable inference methods, such as stochastic gradient Markov chain Monte Carlo (SG-MCMC) that provides posterior samples in a non-batch learning setting (Welling & Teh, 2011; Patterson & Teh, 2013; Ma et al., 2015). Unfortunately, most deep LVMs, such as deep belief network (DBN) (Hinton et al., 2006) and deep Boltzmann machines (DBM) (Salakhutdinov & Hinton, 2009), use greedy layerwise training, without a principled way to jointly learn multilayers in an unsupervised manner (Bengio et al., 2007). While SG-MCMC has recently been successfully applied to several "shallow" LVMs, such as mixture models (Welling & Teh, 2011) and mixed-membership models (Patterson & Teh, 2013), it has been rarely applied to "deep" ones, probably due to the lack of understanding on how to jointly learn the latent variables of different layers and adjust the layer and topic specific learning rates in a non-batch learning setting.

To investigate scalable SG-MCMC inference for deep LVMs, we focus our study on the recently proposed Poisson gamma belief network (PGBN), whose hidden layers are parameterized with gamma distributed hidden units and connected with Dirichlet distributed basis vectors (Zhou et al., 2016a). The PGBN is capable of extracting topics from a text corpus at multiple layers and outperforms a large number of topic modeling algorithms. However, the PGBN is currently trained with a batch Gibbs sampler that is not scalable to big data. In this paper, we focus on developing scalable multilayer joint inference for the PGBN.

We will show that scalable multilayer joint inference of the PGBN could be facilitated by its Fisher information matrix (FIM) (Amari, 1998; Girolami & Calderhead, 2011; Pascanu & Bengio, 2013), which, although seemingly impossible to derive and challenging to work with due to the need to compute the expectations over trigamma functions, is readily available under an alternative representation of the PGBN, referred to as deep latent Dirichlet allocation (DLDA). DLDA, derived by exploiting data augmentation and marginalization techniques on the PGBN, can be considered as a multilayer generalization of latent Dirichlet allocation (LDA) (Blei et al., 2003). Following a general framework for SG-MCMC (Ma et al., 2015), the block diagonal structure of the FIM of DLDA makes it be easily inverted to precondition the mini-batch based noisy gradients to exploit the second-order local curvature information, leading to topic-layer-adaptive step sizes based on the Riemannian manifold and the same asymptotic performance as a natural gradient based batch-learning algorithm (Amari, 1998; Pascanu & Bengio, 2013). To the best of our knowl-edge, this is the first time that the FIM of a deep LVM is shown to have an analytical and practical form. How we derive the FIM for the PGBN using data augmentation and marginalization techniques in this paper may serve as an example to help derive the FIMs for other deep LVMs.

[1]National Laboratory of Radar Signal Processing, Collaborative Innovation Center of Information Sensing and Understanding, Xidian University, Xi'an, China. [2]McCombs School of Business, The University of Texas at Austin, Austin, TX 78712, USA. Correspondence to: Bo Chen <bchen@mail.xidian.edu.cn>, Mingyuan Zhou <mingyuan.zhou@mccombs.utexas.edu>.

Besides presenting the analytical FIM of the PGBN, important for the marriage of a deep LVM and SG-MCMC, we make another contribution in showing how to facilitate SG-MCMC for an LVM equipped with simplex-constrained model parameters $\phi_k = (\phi_{1k}, \ldots, \phi_{Vk})^T$, which means $\sum_{v=1}^{V} \phi_{vk} = 1$ and $\phi_{vk} \in \mathbb{R}_+$, where $\mathbb{R}_+ := \{x, x \geq 0\}$, by using a reduced-mean simplex parameterization together with a fast sampling procedure recently introduced in Cong et al. (2017). Unlike other simplex parameterizations, the reduced-mean one does not make heuristic pseudolikelihood assumptions. Though it has previously been deemed unsound, it is successfully integrated into our SG-MCMC framework to deliver state-of-the-art results. Exploiting the analytical FIM of DLDA and novel inference for simplex-constrained parameters under a general SG-MCMC framework (Ma et al., 2015), we present topic-layer-adaptive stochastic gradient Riemannian (TLASGR) MCMC for DLDA, which automatically adjusts the learning rates of global model parameters across all layers and topics, without the need to set the same learning rate for all that is commonly used in practice due to the difficulty in identifying an appropriate combination of the learning rates for different layers and topics.

## 2. PGBN and SG-MCMC

The generative model of the Poisson gamma belief network (PGBN) (Zhou et al., 2016a) with $L$ hidden layers, from top to bottom, is expressed as

$$\theta_j^{(L)} \sim \text{Gam}\left(r, 1/c_j^{(L+1)}\right),$$
$$\cdots$$
$$\theta_j^{(l)} \sim \text{Gam}\left(\Phi^{(l+1)}\theta_j^{(l+1)}, 1/c_j^{(l+1)}\right), \quad (1)$$
$$\cdots$$
$$x_j^{(1)} \sim \text{Pois}\left(\Phi^{(1)}\theta_j^{(1)}\right), \theta_j^{(1)} \sim \text{Gam}\left(\Phi^{(2)}\theta_j^{(2)}, \frac{p_j^{(2)}}{1-p_j^{(2)}}\right),$$

where the $j^{\text{th}}$ observed or latent $V$-dimensional count vectors $x_j^{(1)} \in \mathbb{Z}^V$, where $\mathbb{Z} := \{0, 1, \ldots\}$, are factorized under the Poisson (Pois) likelihood; the hidden units $\theta_j^{(l)} \in \mathbb{R}_+^{K_l}$ of layer $l$ are factorized under the gamma (Gam) likelihood into the product of the basis vector matrix $\Phi^{(l)} = (\phi_1^{(l)}, \ldots, \phi_{K_l}^{(l)}) \in \mathbb{R}_+^{K_{l-1} \times K_l}$ and the hidden units of the next layer, where $\phi_k^{(l)} \sim \text{Dir}\left(\eta^{(l)}\mathbf{1}_{K_{l-1}}\right)$ are Dirichlet (Dir) distributed and $\mathbf{1}_{K_{l-1}}$ is a $K_{l-1}$-dimensional vector of all ones; the gamma shape parameters $r = (r_1, \cdots, r_{K_L})^T$ at the top layer are shared across all $j$; $\{1/c_j^{(l)}\}_{3,L+1}$ are gamma scale parameters, where $c_j^{(l)} \sim \text{Gam}(e_0, 1/f_0)$, and $c_j^{(2)} := (1-p_j^{(2)})/p_j^{(2)}$, where $p_j^{(2)} \sim \text{Beta}(a_0, b_0)$ are introduced to help reduce the dependencies between $\theta_{jk}^{(1)}$ and $c_j^{(2)}$. The PGBN in (1) can be further extended under the Bernoulli-Poisson link as $b_j^{(1)} = \mathbf{1}(x_j^{(1)} > 0)$, and under the Poisson randomized gamma link as $y_j^{(1)} \sim \text{Gam}(x_j^{(1)}, 1/a_j)$, where $a_j \sim \text{Gam}(e_0, 1/f_0)$.

The PGBN infers a multilayer deep representation of the data, whose inferred basis vectors $\phi_k^{(l)}$ at hidden layer $l$ can be directly visualized as $\left[\prod_{t=1}^{l-1} \Phi^{(t)}\right] \phi_k^{(l)}$, which are their projections into the $V$-dimensional probability simplex. The information of the whole data set is compressed by the PGBN into the inferred sparse network $\{\Phi^{(1)}, \ldots, \Phi^{(L)}\}$, where $\phi_{k_1 k_2}^{(l)}$ indicates the connection strength between node (basis vector) $k_1$ of layer $l-1$ and node $k_2$ of layer $l$. Moreover, the network structure can be inferred from the data by combining the gamma-negative binomial process of Zhou & Carin (2015) with a greedy layer-wise training strategy. Extensive experiments in Zhou et al. (2016a) show that the PGBN can extract basis vectors that are very specific/abstract in the bottom layer and become increasingly more general when moving upwards from the bottom to top hidden layers, and the $K_1$ hidden units $\theta_j^{(1)}$ in the first hidden layer, which are unsupervisedly extracted and regularized with the deep network, are well suited for out-of-sample prediction and being used as features for classification.

Despite all these attractive model properties, the current inference of the PGBN relies on an upward-downward Gibbs sampler that requires processing all data in each iteration and hence often does not scale well to big data unless with parallel computing. To make its inference scalable to allow processing a large amount of data sufficiently fast on a regular personal computer, we resort to SG-MCMC that subsamples the data and utilizes stochastic gradients in each MCMC iteration to generate posterior samples for globally shared model parameters. Let us denote the posterior of model parameters $z$ given the data $X = \{x_j\}_{1,J}$ as $p(z|X) \propto e^{-H(z)}$, with potential function $H(z) = -\ln p(z) - \sum_j \ln p(x_j|z)$. As in Theorem 1 of Ma et al. (2015), $p(z|X)$ is the stationary distribution of the dynamics defined by the stochastic differential equation (SDE) $dz = f(z) dt + \sqrt{2\mathbf{D}(z)}d\mathbf{W}(t)$, if the deterministic drift $f(z)$ is restricted to the form

$$f(z) = -[\mathbf{D}(z) + \mathbf{Q}(z)]\nabla H(z) + \Gamma(z), \quad (2)$$
$$\Gamma_i(z) = \sum_j \frac{\partial}{\partial z_j}[\mathbf{D}_{ij}(z) + \mathbf{Q}_{ij}(z)], \quad (3)$$

where $\mathbf{D}(z)$ is a positive semidefinite diffusion matrix, $\mathbf{W}(t)$ is a Wiener process, $\mathbf{Q}(z)$ is a skew-symmetric curl matrix, and $\Gamma_i(z)$ is the $i$th element of the compensation vector $\Gamma(z)$. Thus one has a mini-batch update rule as

$$z_{t+1} = z_t + \varepsilon_t \left\{ -[\mathbf{D}(z_t) + \mathbf{Q}(z_t)]\nabla \tilde{H}(z_t) + \Gamma(z_t) \right\}$$
$$+ \mathcal{N}\left(\mathbf{0}, \varepsilon_t[2\mathbf{D}(z_t) - \varepsilon_t \hat{\mathbf{B}}_t]\right), \quad (4)$$

where $\varepsilon_t$ denotes step sizes, $\tilde{H}(z) = -\ln p(z) - \rho \sum_{x \in \tilde{X}} \ln p(x|z)$, $\tilde{X}$ the mini-batch, $\rho$ the ratio of the dataset size $|X|$ to the mini-batch size $|\tilde{X}|$, and $\hat{B}_t$ an estimate of the stochastic gradient noise variance satisfying a positive definite constraint as $2D(z_t) - \varepsilon_t \hat{B}_t \succ 0$.

As shown in Ma et al. (2015), stochastic gradient Riemannian Langevin dynamics (SGRLD) of Patterson & Teh (2013) is a special case with $D(z) = G(z)^{-1}, Q(z) = 0, \hat{B}_t = 0$, where $G(z)$ denotes the Fisher information matrix (FIM). SGRLD is designed to solve the inference on the probability simplex, where four different parameterizations of the simplex-constrained basis vectors are discussed, including reduced-mean, expanded-mean, reduced-natural, and expanded-natural. Here, we consider both expanded-mean, previously shown to provide the best overall results, and reduced-mean, which, although discarded in Patterson & Teh (2013) due to its unstable gradients, is used in this paper to produce state-of-the-art results.

Let us denote $\phi_k \in \mathbb{R}_+^V$ as a vector on the probability simplex, $\hat{\phi}_k \in \mathbb{R}_+^V$ as a nonnegative vector, and $\varphi_k \in \mathbb{R}_+^{V-1}$ as a nonnegative vector constrained with $\varphi_{\cdot k} := \sum_{v=1}^{V-1} \varphi_{vk} \leq 1$. For convenience, the symbol "$\cdot$" will denote the operation of summing over the corresponding index. We use $(\hat{\phi}_{1k}, \cdots, \hat{\phi}_{Vk})^T / \sum_v \hat{\phi}_{vk}$ as an expanded-mean parameterization of $\phi_k$ and $(\varphi_{1k}, \cdots, \varphi_{(V-1)k}, 1 - \sum_{v<V} \varphi_{vk})^T$ as a reduced-mean parametrization of $\phi_k$. SGRLD focuses on a single-layer model with a multinomial likelihood $n_k \sim \text{Mult}(n_{\cdot k}, \phi_k)$ and a Dirichlet distributed prior $\phi_k \sim \text{Dir}(\eta \mathbf{1}_V)$. For inference, it adopts the expanded-mean parameterization of $\phi_k$ and makes a heuristic assumption that $n_{\cdot k} \sim \text{Pois}(\hat{\phi}_{\cdot k})$. While that heuristic pseudolikelihood assumption of SGRLD is neither supported by the original generative model nor rigorously justified in theory, it converts a Dirichlet-multinomial model into a gamma-Poisson one, allowing a simple sampling equation for $\hat{\phi}_k$ as

$$(\hat{\phi}_k)_{t+1} = \left| (\hat{\phi}_k)_t + \varepsilon_t \left[ (n_k + \eta) - (n_{\cdot k} + \hat{\phi}_{\cdot k})(\phi_k)_t \right] \right. \\ \left. + \mathcal{N}\left(0, 2\varepsilon_t \text{diag}\left[ (\hat{\phi}_k)_t \right]\right) \right|, \quad (5)$$

where the absolute operation $|\cdot|$ is used to ensure positive-valued $\hat{\phi}_k$. Below we show how to eliminate that heuristic assumption by parameterizing $\phi_k$ with reduced-mean, and develop efficient SG-MCMC for the PGBN, which reduces to LDA when the number of hidden layers is one.

## 3. Deep Latent Dirichlet Allocation

While the original construction of PGBN in (1) makes it seemingly impossible to compute the FIM, as shown in Appendix A, we find that, by exploiting data augmentation and marginalization techniques, the PGBN generative model can be rewritten under an alternative representation that marginalizes out all the gamma distributed hidden units, as shown in the following Lemma, where $\text{Log}(\cdot)$ denotes the logarithmic distribution (Johnson et al., 1997), $m \sim \text{SumLog}(x, p)$ represents the sum-logarithmic distribution generated with $m = \sum_{i=1}^x u_i, u_i \sim \text{Log}(p)$ (Zhou et al., 2016b). The proof is deferred to the Appendix.

**Lemma 3.1.** *Denote* $q_j^{(l+1)} = \ln\left(1 + q_j^{(l)}/c_j^{(l+1)}\right)$ *for* $l = 1, \ldots, L$, *where* $q_j^{(1)} := 1$, *which means* $q_j^{(l+1)} = \ln\left(1 + \frac{1}{c_j^{(l+1)}} \ln\left\{1 + \frac{1}{c_j^{(l)}} \ln\left[1 + \cdots \ln\left(1 + \frac{1}{c_j^{(2)}}\right)\right]\right\}\right)$. *With* $p_j^{(l)} := 1 - e^{-q_j^{(l)}}$ *and* $\tilde{p} := q_{\cdot}^{(L+1)}/(c_0 + q_{\cdot}^{(L+1)})$, *one may re-express the hierarchical model of the PGBN as deep latent Dirichlet allocation (DLDA) as*

$$x_{k\cdot}^{(L+1)} \sim \text{Log}(\tilde{p}), \quad K_L \sim \text{Pois}[-\gamma_0 \ln(1-\tilde{p})],$$
$$X^{(L+1)} = \sum_{k=1}^{K_L} x_{k\cdot}^{(L+1)} \delta_{\phi_k^{(L)}},$$
$$(x_{vj}^{(L+1)})_j \sim \text{Mult}\left[x_{v\cdot}^{(L+1)}, (q_j^{(L+1)})_j / q_{\cdot}^{(L+1)}\right],$$
$$m_{vj}^{(L)(L+1)} \sim \text{SumLog}(x_{vj}^{(L+1)}, p_j^{(L+1)}),$$
$$\cdots$$
$$x_{vj}^{(l)} = \sum_{k=1}^{K_l} x_{vkj}^{(l)}, \quad (x_{vkj}^{(l)})_v \sim \text{Mult}\left(m_{kj}^{(l)(l+1)}, \phi_k^{(l)}\right),$$
$$m_{vj}^{(l-1)(l)} \sim \text{SumLog}(x_{vj}^{(l)}, p_j^{(l)}),$$
$$\cdots$$
$$x_{vj}^{(1)} = \sum_{k=1}^{K_1} x_{vkj}^{(1)}, \quad (x_{vkj}^{(1)})_v \sim \text{Mult}\left(m_{kj}^{(1)(2)}, \phi_k^{(1)}\right). \quad (6)$$

Note that the equations in the first four lines of (6) precisely represent a random count matrix generated from a gamma-negative binomial process that can also be generated from

$$x_{kj}^{(L+1)} \sim \text{Pois}\left(r_k q_j^{(L+1)}\right), \quad r_k \sim \text{Gam}(\gamma_0/K, 1/c_0),$$
$$m_{kj}^{(L)(L+1)} \sim \text{SumLog}\left(x_{kj}^{(L+1)}, p_j^{(L+1)}\right) \quad (7)$$

by letting $K \to \infty$ (Zhou et al., 2016b). When $L = 1$, the PGBN whose $(r_k, \phi_k)$ are the points of a gamma process reduces to the gamma-negative binomial process PFA of Zhou & Carin (2015), whose alternative representation is provided in Corollary D.1 in the Appendix. Note that how we re-express the PGBN as DLDA is related to how Schein et al. (2016) re-express their Poisson–gamma dynamic systems into an alternative representation that facilitates inference.

DLDA, designed to infer a multilayer representation of observed or latent high-dimensional sparse count vectors, constrains all the basis vectors of different layers to probability simplices. It is clear from (6) that a data point backpropagates its counts through the network one layer at a time via a sum-logarithmic distribution to *enlarge* each element of a $K_l$-dimensional count vector, a multinomial distribution to *partition* that enlarged count vector into a $K_{l-1} \times K_l$ count matrix, and then a row-sum operation to *aggregate* that latent count matrix into a $K_{l-1}$-dimensional count vector, where $K_0 := V$ is the feature dimension. Below we show that such an alternative representation that repeats the *enlarge-partition-augment* operation brings significant benefits when it comes to deriving SG-MCMC inference with preconditioned gradients.

### 3.1. Fisher Information Matrix of DLDA

In deep LVMs, whose parameters of different layers are often highly correlated to each other, it is often difficult to tune the step sizes of different layers together and hence one often chooses to train an unsupervised deep model in a greedy layer-wise manner (Bengio et al., 2007), which is a sensible but not optimal training strategy. To address that issue, we resort to the inverse of the FIM that is widely used to precondition the gradients to adjust the learning rates of different model parameters (Amari, 1998; Pascanu & Bengio, 2013; Ma et al., 2015; Li et al., 2016). However, it is often difficult to compute the FIMs of deep LVMs as

$$\mathbf{G}(\mathbf{z}) = \mathbb{E}_{\mathbf{\Omega}|\mathbf{z}}\left[-\frac{\partial^2}{\partial \mathbf{z}^2}\ln p(\mathbf{\Omega}|\mathbf{z})\right], \quad (8)$$

where $\mathbf{z}$ denotes the set of all global variables and $\mathbf{\Omega}$ is the set of all observed and local variables.

Although deriving the FIM for the PGBN generative model shown in (1) seems impossible, we find it to be straightforward under the alternative DLDA representation in (6). Since the likelihood of (6) is fully factorized with respect to the global parameters $\mathbf{z}$, *i.e.*, $\boldsymbol{\phi}_k^{(l)}$ and $\boldsymbol{r}$, one may readily show the FIM $\mathbf{G}(\mathbf{z})$ of (6) has a block diagonal form as

$$\text{diag}\left[\mathbf{I}\left(\boldsymbol{\varphi}_1^{(1)}\right), \cdots, \mathbf{I}\left(\boldsymbol{\varphi}_{K_L}^{(L)}\right), \mathbf{I}(\boldsymbol{r})\right]; \quad (9)$$

with the likelihood $\left(x_{vkj}^{(l)}\right)_v \sim \text{Mult}\left(m_{kj}^{(l)(l+1)}, \boldsymbol{\phi}_k^{(l)}\right)$ and the reduced-mean parameterization, we have

$$\mathbf{I}\left(\boldsymbol{\varphi}_k^{(l)}\right) = -\mathbb{E}\left[\frac{\partial^2}{\partial \boldsymbol{\varphi}_k^{(l)2}}\ln\left(\prod_j \text{Mult}\left[(x_{vkj}^{(l)})_v; m_{kj}^{(l)(l+1)}, \boldsymbol{\phi}_k^{(l)}\right]\right)\right]$$

$$= M_k^{(l)}\left[\text{diag}\left(1/\boldsymbol{\varphi}_k^{(l)}\right) + \mathbf{1}\mathbf{1}^T/(1-\varphi_{\cdot k}^{(l)})\right], \quad (10)$$

where $M_k^{(l)} := \mathbb{E}\left[m_{k\cdot}^{(l)(l+1)}\right] = \mathbb{E}\left[x_{\cdot k\cdot}^{(l)}\right]$. Similarly, with the likelihood $x_{kj}^{(L+1)} \sim \text{Pois}(r_k q_j^{(L+1)})$, we have

$$\mathbf{I}(\boldsymbol{r}) = M^{(L+1)}\text{diag}(1/\boldsymbol{r}), \quad (11)$$

where $M^{(L+1)} := \mathbb{E}\left[q_\cdot^{(L+1)}\right]$.

The block diagonal structure of the FIM of DLDA makes it computationally appealing to apply its inverse for preconditioning. Under the framework suggested by (4), we adopt the similar settings used in SGRLD (Patterson & Teh, 2013) that lets $\mathbf{D}(\mathbf{z}) = \mathbf{G}(\mathbf{z})^{-1}$, $\mathbf{Q}(\mathbf{z}) = \mathbf{0}$, and $\hat{\mathbf{B}}_t = \mathbf{0}$. While other more sophisticated settings described in Ma et al. (2015), including as special examples stochastic gradient Hamiltonian Monte Carlo in Chen et al. (2014) and stochastic gradient thermostats in Ding et al. (2014), may be used to further improve the performance, we choose this specific one to make a direct comparison with SGRLD.

By substituting the FIM $\mathbf{G}(\mathbf{z})$ and the adopted settings into (4), it is apparent that we only need to choose a single step size $\varepsilon_t$, relying on the FIM to automatically adjust the relatively learning rates for different parameters across all layers and topics. Moreover, the block-diagonal structure of $\mathbf{G}(\mathbf{z})$ will be carried over to its inverse $\mathbf{D}(\mathbf{z})$, making it simple to perform updating using (4), as described below.

### 3.2. Inference on the Probability Simplex

As discussed in Section 2, to sample simplex-constrained model parameters for a Dirichlet-multinomial model, the SGRLD of Patterson & Teh (2013) adopts the expanded-mean parameterization of simplex-constrained vectors and makes a pseudolikelihood assumption to simplify the derivation of update equations. In this paper, without replying on that pseudolikelihood assumption, we choose to use the reduced-mean parameterization of simplex-constrained vectors, despite being considered as an unsound choice in Patterson & Teh (2013). In the following discussion, we omit the layer-index superscript $(l)$ for simplicity.

With the multinomial likelihood in (6) and the Dirichlet-multinomial conjugacy, the conditional posterior of $\phi_k$ can be expressed as $(\phi_k | -) \sim \text{Dir}(x_{1k\cdot} + \eta, \ldots, x_{Vk\cdot} + \eta)$. Taking the gradient with respect to $\boldsymbol{\varphi}_k \in \mathbb{R}_+^{V-1}$ on the summation of the negative log-likelihood of a mini-batch $\tilde{X}$ scaled by $\rho = |X|/|\tilde{X}|$ and the negative log-likelihood of the Dirichlet prior, we have

$$\nabla_{\boldsymbol{\varphi}_k}\left[-\tilde{H}(\boldsymbol{\varphi}_k)\right] = \frac{\rho\bar{\boldsymbol{x}}_{:k\cdot} + \eta - 1}{\boldsymbol{\varphi}_k} - \frac{\rho\tilde{x}_{Vk\cdot} + \eta - 1}{1 - \varphi_{\cdot k}}, \quad (12)$$

where $\tilde{x}_{vk\cdot} = \sum_{j:x_j \in \tilde{X}} x_{vkj}$ and $\bar{\boldsymbol{x}}_{:k\cdot} := (\tilde{x}_{1k\cdot}, \ldots, \tilde{x}_{(V-1)k\cdot})^T$. Note the gradient in (12) becomes unstable when some components of $\boldsymbol{\varphi}_k$ approach zeros, a key reason that this approach is mentioned but not further pursued in Patterson & Teh (2013).

However, after preconditioning the noisy gradient with the inverse of the FIM, it is intriguing to find out that the stability issue completely disappears. More specifically, by plugging both the FIM in (10) and noisy gradient in (12) into the SG-MCMC update in (4), a noisy estimate of the deterministic drift defined in (2) obtained using the current mini-batch can be expressed as

$$\mathbf{I}(\boldsymbol{\varphi}_k)^{-1}\nabla_{\boldsymbol{\varphi}_k}\left[-\tilde{H}(\boldsymbol{\varphi}_k)\right] + \Gamma(\boldsymbol{\varphi}_k)$$
$$= M_k^{-1}\left[(\rho\bar{\boldsymbol{x}}_{:k\cdot} + \eta) - (\rho\tilde{x}_{\cdot k\cdot} + \eta V)\boldsymbol{\varphi}_k\right], \quad (13)$$

where $\Gamma(\boldsymbol{\varphi}_k) = M_k^{-1}[1 - V\boldsymbol{\varphi}_k]$ according to (3), as derived in detail in Appendix B. Consequently, with $[\cdot]_\triangle$ de-noting the constraint that $\varphi_{vk} \geq 0$ and $\sum_{v=1}^{V-1} \varphi_{vk} \leq 1$, using (4), the sampling of $\boldsymbol{\varphi}_k$ becomes

$$(\boldsymbol{\varphi}_k)_{t+1} = \left[(\boldsymbol{\varphi}_k)_t + \frac{\varepsilon_t}{M_k}\left[(\rho\bar{\boldsymbol{x}}_{:k\cdot} + \eta) - (\rho\tilde{x}_{\cdot k\cdot} + \eta V)(\boldsymbol{\varphi}_k)_t\right]\right.$$
$$\left. + \mathcal{N}\left(\mathbf{0}, \frac{2\varepsilon_t}{M_k}\left[\text{diag}\,(\boldsymbol{\varphi}_k)_t - (\boldsymbol{\varphi}_k)_t(\boldsymbol{\varphi}_k)_t^T\right]\right)\right]_\triangle. \quad (14)$$

Even without the $[\cdot]_\triangle$ constraint, the multivariate normal (MVN) simulation in (14), although easy to interpret and numerically stable, is computationally expensive if the Cholesky decomposition, with $\mathcal{O}((V-1)^3)$ complexity (Golub & Van Loan, 2012), is adopted directly. Fortunately, using Theorem 2 of Cong et al. (2017), the special structure of its covariance matrix allows an equivalent but substantially more efficient simulation of $\mathcal{O}(V)$ complexity by transforming a random variable drawn from a related MVN that has a diagonal covariance matrix. More specifically, the sampling of (14) can be efficiently realized in a $V$-dimensional space as

$$(\boldsymbol{\phi}_k)_{t+1} = \left[(\boldsymbol{\phi}_k)_t + \frac{\varepsilon_t}{M_k}\left[(\rho\tilde{\boldsymbol{x}}_{:k\cdot} + \eta) - (\rho\tilde{x}_{\cdot k\cdot} + \eta V)(\boldsymbol{\phi}_k)_t\right]\right.$$
$$\left. + \mathcal{N}\left(\mathbf{0}, \frac{2\varepsilon_t}{M_k}\text{diag}\,(\boldsymbol{\phi}_k)_t\right)\right]_\angle, \quad (15)$$

where $[\cdot]_\angle$ denotes the simplex constraint that $\phi_{vk} \geq 0$ and $\sum_{v=1}^V \phi_{vk} = 1$. More details on simulating (14) and (15) can be found in Examples 1-3 of Cong et al. (2017).

Similarly, with the gamma-Poisson construction in (7), we have $\Gamma_k(\boldsymbol{r}) = 1/M^{(L+1)}$, as in Appendix B, and

$$\nabla_{r_k}\left[-\tilde{H}(\boldsymbol{r})\right] = r_k^{-1}\left(\rho\tilde{x}_{k\cdot}^{(L+1)} + \frac{\gamma_0}{K_L} - 1\right) - \left(c_0 + \rho\tilde{q}_{\cdot}^{(L+1)}\right), \quad (16)$$

which also becomes unstable if $r_k$ approaches zero. Substituting (16) and (11) into (4) leads to

$$\boldsymbol{r}_{t+1} = \left|\boldsymbol{r}_t + \frac{\varepsilon_t}{M^{(L+1)}}\left[\left(\rho\tilde{\boldsymbol{x}}_{:\cdot}^{(L+1)} + \frac{\gamma_0}{K_L}\right) - \boldsymbol{r}_t\left(c_0 + \rho\tilde{q}_{\cdot}^{(L+1)}\right)\right]\right.$$
$$\left. + \mathcal{N}\left(\mathbf{0}, \frac{2\varepsilon_t}{M^{(L+1)}}\text{diag}\,(\boldsymbol{r}_t)\right)\right|, \quad (17)$$

for which there is no stability issue.

### 3.3. Topic-Layer-Adaptive Stochastic Gradient Riemannian MCMC

Note that $M^{(L+1)}$ and $M_k^{(l)}$ for $l \in \{1, \ldots, L\}$, appearing as denominates in (17) and (15), respectively, are expectations that need to be approximately calculated. We update them using annealed weighting (Polatkan et al., 2015) as

$$M_k^{(l)} = \left(1 - \varepsilon_t'\right)M_k^{(l)} + \varepsilon_t'\rho E\left[\tilde{x}_{\cdot k\cdot}^{(l)}\right], \quad (18)$$

$$M^{(L+1)} = \left(1 - \varepsilon_t'\right)M^{(L+1)} + \varepsilon_t'\rho E\left[\tilde{q}_{\cdot}^{(L+1)}\right], \quad (19)$$

where $E[\cdot]$ denotes averaging over the collected MCMC samples. For simplicity, we set $\varepsilon_t' = \varepsilon_t$ in this paper, which is found to work well in practice.

**Algorithm 1** TLASGR MCMC for DLDA (PGBN).

**Input:** Data mini-batches;
**Output:** Global parameters of DLDA (PGBN).
1: **for** $t = 1, 2, \cdots$ **do**
2:    /* Collect local information
3:    Upward-downward Gibbs sampling (Zhou et al., 2016a) on the $t^{\text{th}}$ mini-batch for $\tilde{\boldsymbol{x}}_{:k\cdot}, \tilde{x}_{\cdot k\cdot}, \tilde{\boldsymbol{x}}_{:\cdot}^{(L+1)}$, and $\tilde{q}_{\cdot}^{(L+1)}$;
4:    /* Update global parameters
5:    **for** $l = 1, \cdots, L$ and $k = 1, \cdots, K_l$ **do**
6:      Update $M_k^{(l)}$ with (18); then $\boldsymbol{\phi}_k^{(l)}$ with (15);
7:    **end for**
8:    Update $M^{(L+1)}$ with (19) and then $\boldsymbol{r}$ with (17).
9: **end for**

Note that as in (15) and (17), instead of having a single learning rate for all layers and topics, a common practice due to the difficulty to adapt the step sizes to different layers and/or topics, the proposed inference employs topic-layer-adaptive learning rates as $\varepsilon_t/M_k^{(l)}$, where $M_k^{(L+1)} := M^{(L+1)}$, adapting a single step size $\varepsilon_t$ to different topics and layers by multiplying it with the weights $1/M_k^{(l)}$ for $l \in \{1, \ldots, L\}$ and $k \in \{1, \ldots, K_l\}$. We refer to the proposed inference algorithm with adaptive learning rates as topic-layer-adaptive stochastic gradient Riemannian (TLASGR) MCMC, as summarized in Algorithm 1 that is simple to implement.

## 4. Related Work

Both LDA (Blei et al., 2003) and the related Poisson factor analysis (PFA) (Zhou et al., 2012) are equipped with scalable inference, such as stochastic variational inference (SVI) (Hoffman et al., 2010; Mimno et al., 2012) and SGRLD (Patterson & Teh, 2013). However, both are shallow LVMs whose modeling capacities are often insufficient for big and complex data. The deep Poisson factor models of Gan et al. (2015) and Henao et al. (2015) are proposed to generalize PFA with deep structures, but both of them only explore the deep information in binary topic usage patterns instead of the full connection weights that are used in the PGBN. The proposed DLDA shares some similarities with the pachinko allocation model of Li & McCallum (2006) in that they both adopt layered construction and use Dirichlet distributed topics. Ranganath et al. (2015) propose deep exponential family (DEF), which differs from the PGBN in connecting adjacent layers via the gamma rate parameters and using black-box variational inference (BBVI) (Ranganath et al., 2014).

Some commonly used neural networks, such as deep belief network (DBN) (Hinton et al., 2006) and deep Boltzmann machines (DBM) (Salakhutdinov & Hinton, 2009), have also been modified for text analysis (Hinton & Salakhutdinov, 2009; Larochelle & Lauly, 2012; Srivastava et al., 2013). Although they may work well for certain text analysis tasks, they are not naturally designed for count data and often yield latent structures that are not readily interpretable.

The neural variational document model (NVDM) of Miao et al. (2016), even though using deep neural networks in its variational auto-encoder (VAE) (Kingma & Welling, 2013), still relies on a single-layer model for data generalization.

Generally speaking, it is challenging to develop an efficient and principled multilayer joint learning algorithm for deep LVMs. Scalable variational inference, such as BBVI, often makes the restrictive mean-field assumption. Neural variational inference and learning (NVIL) relies on variance reduction techniques that are often difficult to be generalized for discrete LVMs (Mnih & Gregor, 2014; Rezende et al., 2014). When a SG-MCMC algorithm is used, a single learning rate is often applied for different variables across all layers (Welling & Teh, 2011; Neal et al., 2011; Chen et al., 2014; Ding et al., 2014). It is possible to improve SG-MCMC by adjusting its noisy gradients with some stochastic optimization technique, such as Adagrad (Duchi et al., 2011), Adadelta (Zeiler, 2012), Adam (Kingma & Ba, 2014), and RMSprop (Tieleman & Hinton, 2012). For example, Li et al. (2016) show that preconditioning the gradients with diagonal approximated FIM improves SG-MCMC in both training speed and predictive accuracy for supervised learning where gradients are easy to calculate. Other efforts exploiting similar preconditioning idea focus on shallow and/or binary models (Mimno et al., 2012; Patterson & Teh, 2013; Grosse & Salakhutdinov, 2015; Song et al., 2016; Simsekli et al., 2016), and it is unclear how that idea can be extended to deep LVMs whose gradients and FIM maybe difficult to approximate.

## 5. Experiment results

We present experimental results on three benchmark corpora: 20Newsgroups (20News), Reuters Corpus Volume I (RCV1) that is moderately large, and Wikipedia (Wiki) that is huge. 20News consists of 18,845 documents with a vocabulary size of 2,000, partitioned into 11,315 training documents and 7,531 test ones. RCV1 consists of 804,414 documents with a vocabulary size of 10,000, where 10,000 documents are randomly selected for testing. Wiki consists of 10 million documents randomly downloaded from *Wikipedia* using the scripts provided in Hoffman et al. (2010); as in Hoffman et al. (2010), Gan et al. (2015), and Henao et al. (2015), we use a vocabulary with 7,702 words and randomly select 1,000 documents for testing. To make a fair comparison, these corpora, including the training/testing partitions, are set to be the same as those in Gan et al. (2015) and Henao et al. (2015). To be consistent with the settings of Gan et al. (2015) and Henao et al. (2015), no precautions are taken in the scripts for *Wikipedia* to prevent a testing document from being downloaded into a mini-batch for training.

We consider two related performance measures. The first one is the commonly-used per-heldout-word perplexity calculated as follows: for each test document, we randomly select 80% of the word tokens to sample the local variables specific to the document, under the global model parameters of each MCMC iteration; after the burn-in period, we accumulate the first layer's Poisson rates in each collected MCMC sample; in the end, we normalize these accumulated Poisson rates to calculate the perplexity using the remaining 20% word tokens. Similar evaluation methods have been widely used, *e.g.*, in Wallach et al. (2009), Paisley et al. (2011), and Zhou & Carin (2015). Although a good measure for overall performance, the per-heldout-word perplexity, calculated based on multiple collected MCMC samples of global parameters, may not be ideal to check the performance in real time to assess how efficient an iterative algorithm improves its performance as time increases. Therefore, we slightly modify it to provide a *point* per-heldout-word perplexity calculated based on only the global parameters of the most recent MCMC sample. For simplicity, we refer to (point) per-heldout-word perplexity as (point) perplexity.

For comparison, we consider LDA of Blei et al. (2003), focused topic model (FTM) of Williamson et al. (2010), replicated softmax (RSM) of Hinton & Salakhutdinov (2009), nested Hierarchical Dirichlet process (nHDP) of Paisley et al. (2015), DPFA of Gan et al. (2015), and DPFM of Henao et al. (2015). For these methods, the perplexity results are taken from Gan et al. (2015) and Henao et al. (2015). For the proposed algorithms, we set the mini-batch size as 200, and use as burn-in 2000 mini-batches for both 20News and RCV1 and 3500 mini-batches for Wiki. We collect 1500 samples to calculate perplexity. For point perplexity, given the global parameters of an MCMC sample, we sample the local variables with 600 iterations and collect one sample every two iterations during the last 400 iterations. The hyperparameters of DLDA are set as: $\eta^{(l)} = 1/K_l$, $a_0 = b_0 = 0.01$, and $\gamma_0 = c_0 = e_0 = f_0 = 1$. Note $\eta^{(l)}$ and $K_l$ are set similar to that of DPFM for fair comparisons, while other hyperparameters follow Zhou et al. (2016a).

To demonstrate the advantages of using the reduced-mean simplex parameterization and inverting the FIM for preconditioning to obtain topic-layer-adaptive learning rates, we consider four different inference methods:

1) TLASGR: topic-layer-adaptive stochastic gradient Riemannian MCMC for DLDA, as described in Algorithm 1.

2) TLFSGR: topic-layer-fixed stochastic gradient Riemannian MCMC for DLDA that replaces the adaptive learning rates $\varepsilon_t / M_k^{(l)}$ of TLASGR with $\varepsilon_t / (\sum_{k=1}^{K_1} M_k^{(1)} / K_1)$.

3) SGRLD: updating each $\phi_k^{(l)}$ under the expanded-mean parameterization as in (5), served as a good scalable baseline for comparison since it was shown in Patterson & Teh (2013) to perform significantly better than SVI. It differs from TLFSGR mainly in using a different parameterization for

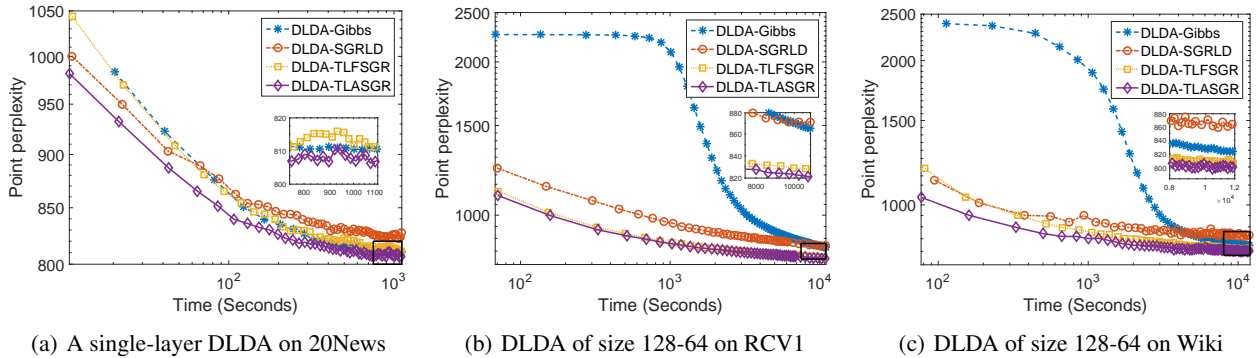

(a) A single-layer DLDA on 20News  (b) DLDA of size 128-64 on RCV1  (c) DLDA of size 128-64 on Wiki

*Figure 1.* Plot of point perplexity as a function of time. (a) 20News with a single-layer DLDA with 128 topics. (b) RCV1 with a two-layer DLDA with 128 and 64 topics in the first and second layers, respectively. (c) Wiki with a two-layer DLDA, with 128 and 64 topics in the first and second layers, respectively. Note a small subset of $10^6$ documents from Wiki is used for demonstration.

*Table 1.* Per-heldout-word perplexities on 20 News, RCV1 and Wiki. For models except DLDA, the results are taken from Gan et al. (2015) and Henao et al. (2015). Note that for Wiki, DPFM with MCMC infers the global parameters on a subset of the corpus with 3,000 MCMC iterations.

| Model | Method | Size | 20 News | RCV1 | Wiki |
|---|---|---|---|---|---|
| DLDA | TLASGR | 128-64-32 | 757 | 815 | **786** |
| DLDA | TLASGR | 128-64 | 758 | 817 | 787 |
| DLDA | TLASGR | 128 | 770 | 823 | 802 |
| DLDA | TLFSGR | 128-64-32 | 760 | 817 | 789 |
| DLDA | TLFSGR | 128-64 | 759 | 819 | 791 |
| DLDA | TLFSGR | 128 | 772 | 829 | 804 |
| DLDA | SGRLD | 128-64-32 | 775 | 827 | 792 |
| DLDA | SGRLD | 128-64 | 770 | 823 | 792 |
| DLDA | SGRLD | 128 | 777 | 829 | 803 |
| DLDA | Gibbs | 128-64-32 | **752** | **802** | — |
| DLDA | Gibbs | 128-64 | 754 | 804 | — |
| DLDA | Gibbs | 128 | 768 | 818 | — |
| DPFM | SVI | 128-64 | 818 | 961 | 791 |
| DPFM | MCMC | 128-64 | 780 | 908 | **783** |
| DPFA-SBN | SGNHT | 128-64-32 | 827 | 1143 | 876 |
| DPFA-RBM | SGNHT | 128-64-32 | 896 | 920 | 942 |
| nHDP | SVI | (10,10,5) | 889 | 1041 | 932 |
| LDA | Gibbs | 128 | 893 | 1179 | 1059 |
| FTM | Gibbs | 128 | 887 | 1155 | 991 |
| RSM | CD5 | 128 | 877 | 1171 | 1001 |

$\phi_k^{(l)}$ and adding a pseudolikelihood assumption.

4) Gibbs: the upward-downward Gibbs sampler in Zhou et al. (2016a).

Both TLASGR and TLFSGR differ from SGRLD mainly in how the global parameters $\phi_k^{(l)}$ are updated. While TLASGR uses topic-layer-adaptive learning rates, both TLFSGR and SGRLD use a single learning rate, a common practice due to the difficulty of tuning the step sizes across layers and topics. We keep the same stepsize schedule of $\varepsilon_t = a(1 + t/b)^{-c}$ as in Patterson & Teh (2013) and Ma et al. (2015).

Let us first examine how various inference algorithms perform on 20News with a single-layer DLDA of size 128, which can be considered as a topic model that imposes an asymmetric prior on a document's proportion over these 128 topics. Under this setting, as shown in Fig. 1(a), TLFSGR clearly outperforms SGRLD in providing lower point perplexities as time progresses, which is not surprising as under the reduced-mean simplex parameterization, to derive its sampling equations, TLFSGR does not rely on a pseudolikelihood assumption that is adopted by SGRLD in its expanded-mean simplex parameterization. Moreover, TLASGR is found to further improve TLFSGR, suggesting that even for a single-layer model, replacing a fixed learning rate as $\varepsilon_t/(\sum_{k=1}^{K_1} M_k^{(1)}/K_1)$ with topic-adaptive learning rates as $\varepsilon_t/M_k^{(1)}$ could further improve the performance.

Let us then examine how these algorithms perform on two larger corpora—RCV1 and Wiki—with a two-layer DLDA of size 128-64, which improves the single-layer one by capturing the co-occurrence patterns between the topics of the first layer with those of the second layer (Zhou et al., 2016a). As show in Figs. 1(b) and 1(c), it is clear that the proposed TLASGR performs the best for the two-layer DLDA and consistently outperforms TLFSGR as time progresses. In comparison, SGRLD quickly improves its performance as a function of time in the beginning but its point perplexity remains higher even after 10,000 seconds, whereas Gibbs sampling slowly improves its performance as a function of time in the beginning but moves its point perplexity closer and closer to that of TLASGR as time progresses.

Note that for 20News, the point perplexity of the mini-batch based TLASGR quickly decreases as time increases, while that of Gibbs sampling decreases relatively slowly. That discrepancy of convergence rate as a function of time becomes much more evident for both RCV1 and Wiki, as shown in Figs. 1(b) and 1(c). This is expected as both RCV1 and Wiki are much larger corpora, for which a mini-batch based inference algorithm can already make significant progress in learning the global model parameters, before a batch-learning Gibbs sampler finishes a single iteration that needs to go through all documents.

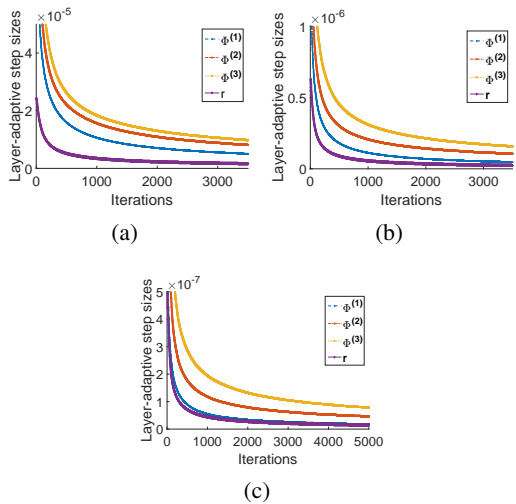

Figure 2. Topic-layer-adaptive learning rates inferred with a three-layer DLDA of size 128-64-32. (a) 20News. (b) RCV1. (c) Wiki. Note the layer-adaptive learning rate for layer $l$ is obtained by averaging over the topic-layer-adaptive learning rates of all $\phi_k^{(l)}$ for $k = 1, \ldots, K_l$.

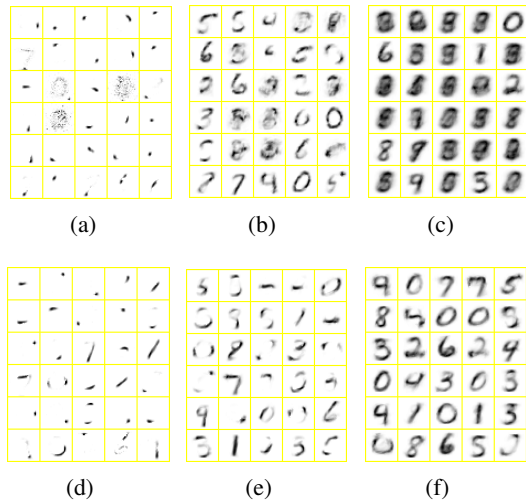

Figure 3. Learned dictionary atoms on MNIST digits with a three-layer GBN of size 128-64-32 after one full epoch. Shown in (a)-(c) are example atoms for $\phi_k^{(1)}$, $\Phi^{(1)}\phi_k^{(2)}$, and $\Phi^{(1)}\Phi^{(2)}\phi_k^{(3)}$, respectively, learned with TLFSGR, and shown in (d)-(f) are example ones learned with TLASGR.

To illustrate the working mechanism of TLASGR, we show how its inferred learning rates are adapted to different layers in Fig. 2. By contrast, TLFSGR admits a fixed learning rate that leads to worse performance. Several interesting observations can be made for TLASGR from Figs. 2(a)-2(c): 1) for $\Phi^{(l)}$, higher layers prefer larger step sizes, which may be explained by the *enlarge-partition-augment* data generating mechanism of DLDA; 2) larger datasets prefer slower learning rates, reflected by the scales of the vertical axes; 3) and the relative learning rates between different layers vary across different datasets.

To further verify the excellent performance of DLDA inferred with TLASGR, we compare a wide variety of models and inference algorithms in Table 1. For 20News and RCV1, DLDA with Gibbs sampling performs the best in terms of perplexity and exhibits a clear trend of improvement as the number of hidden layers increases. For Wiki, a single iteration of the DLDA Gibbs sampler on the full corpus is so expensive in both time and memory that its performance is not reported. For DLDA on 20News and RCV1, TLASGR only slightly underperforms Gibbs sampling, and the performance degradation from Gibbs sampling to TLASGR is significantly smaller than that from MCMC to SVI for DPFM. That relative small degradation caused by changing from Gibbs sampling to the mini-batch based TLASGR could be attributed to the Fisher efficiency brought by the FIM. Generally speaking, in comparison to SGRLD, TLASGR brings a clear boost in performance, which is particularly evident for a deeper DLDA, and TLASGR consistently outperforms TLFSGR that does not adapt its learning rates to different topics and layers.

**MNIST.** To further illustrate the advantages of using the inverse of the FIM for preconditioning in a deep generative model, and to visualize the benefits of automatically adjusting the relative learning rates of different hidden layers, we apply a three-layer Poisson randomized gamma gamma belief network (PRG-GBN) (Zhou et al., 2016a) to 60,000 MNIST digits and present the learned dictionary atoms after one full epoch, as shown in Fig. 3. It is clear that, with topic-layer-adaptive learning rates, which are made possible by utilizing the FIM, TLASGR provides more effective mini-batch based stochastic updates to allow better information propagation between different hidden layers, extracting more interpretable features at multiple layers.

## 6. Conclusions

For scalable multilayer joint inference of the Poisson gamma belief network (PGBN), we introduce an alternative representation of the PGBN, which is referred to as deep latent Dirichlet allocation (DLDA) that can be considered as a multilayer generalization of latent Dirichlet allocation. We show how to reparameterize the simplex constrained basis vectors, derive a block-diagonal Fisher information matrix (FIM), and efficiently compute the inverse of the FIM, leading to a stochastic gradient MCMC algorithm referred to as topic-layer-adaptive stochastic gradient Riemannian (TLASGR) MCMC. The proposed TLASGR-MCMC is able to jointly learn the parameters of different layers with topic-layer-adaptive step sizes, which makes DLDA (PGBN) much more practical in a big data setting. Compelling experimental results on large text corpora and the MNIST dataset demonstrated the advantages of TLASGR-MCMC.

# Acknowledgements

Bo Chen thanks the support of the Thousand Young Talent Program of China, NSFC (61372132), and NDPR-9140A07010115DZ01019. Hongwei Liu thanks the support of NSFC for Distinguished Young Scholars (61525105).

# Supplementary Material for Deep latent Dirichlet allocation with topic-layer-adaptive stochastic gradient Riemannian MCMC

Yulai Cong, Bo Chen, Hongwei Liu, and Mingyuan Zhou

## A. Naive derivation of the Fisher information matrix of the Poisson gamma belief network

For simplicity, we take for example a two-layer Poisson gamma belief network (PGBN), expressed as

$$\boldsymbol{\theta}_j^{(2)} \sim \text{Gam}\left(\boldsymbol{r}, 1/c_j^{(3)}\right),$$
$$\boldsymbol{x}_j^{(1)} \sim \text{Pois}\left(\boldsymbol{\Phi}^{(1)}\boldsymbol{\theta}_j^{(1)}\right), \boldsymbol{\theta}_j^{(1)} \sim \text{Gam}\left(\boldsymbol{\Phi}^{(2)}\boldsymbol{\theta}_j^{(2)}, \frac{p_j^{(2)}}{1-p_j^{(2)}}\right), \quad (20)$$

and focus on a specific element $\boldsymbol{\Phi}_{vk}^{(2)}$ only.

With the definition in (8), it is straight to show that the $\boldsymbol{\Phi}^{(2)}$-relevant part in $\ln p(\boldsymbol{\Omega}|\boldsymbol{z})$ is

$$\sum_{vj}\left[\boldsymbol{\Phi}_{v:}^{(2)}\boldsymbol{\theta}_{:j}^{(2)}\ln\left(c_j^{(2)}\boldsymbol{\theta}_{vj}^{(1)}\right) - \ln\Gamma\left(\boldsymbol{\Phi}_{v:}^{(2)}\boldsymbol{\theta}_{:j}^{(2)}\right)\right]. \quad (21)$$

Accordingly, for $\boldsymbol{\Phi}_{vk}^{(2)}$, we have

$$\mathbb{E}\left[-\frac{\partial^2}{\partial[\boldsymbol{\Phi}_{vk}^{(2)}]^2}\ln p(\boldsymbol{\Omega}|\boldsymbol{z})\right] = \mathbb{E}\left[\sum_j \psi'\left(\boldsymbol{\Phi}_{v:}^{(2)}\boldsymbol{\theta}_{:j}^{(2)}\right)\left[\boldsymbol{\theta}_{:j}^{(2)}\right]^2\right], \quad (22)$$

where $\psi'(\cdot)$ is the trigamma function. This expectation involving the trigamma function is difficult to calculate.

## B. Derivation of the $\Gamma(\cdot)$ functions in Section 3.2

With $\mathbf{D}(\boldsymbol{z}) = \mathbf{G}(\boldsymbol{z})^{-1}$, $\mathbf{Q}(\boldsymbol{z}) = \mathbf{0}$, and the block-diagonal Fisher information matrix (FIM) $\mathbf{G}(\boldsymbol{z})$ in (9), it is straight to show that $\frac{\partial}{\partial \boldsymbol{\varphi}_k}[\mathbf{D}(\boldsymbol{z}) + \mathbf{Q}(\boldsymbol{z})]$ is non-zero only in the $\boldsymbol{\varphi}_k$-related block $\mathbf{I}(\boldsymbol{\varphi}_k)$ in (10). Therefore, we focus on this block and have

$$\Gamma_v(\boldsymbol{\varphi}_k) = \sum_u \frac{\partial}{\partial \varphi_{uk}}\left[\mathbf{I}_{vu}^{-1}(\boldsymbol{\varphi}_k)\right], \quad (23)$$

where $\mathbf{I}^{-1}(\boldsymbol{\varphi}_k) = M_k^{-1}\left[\text{diag}(\boldsymbol{\varphi}) - \boldsymbol{\varphi}\boldsymbol{\varphi}^T\right]$. Accordingly, we have

$$\Gamma_v(\boldsymbol{\varphi}_k) = M_k^{-1}\sum_u \frac{\partial}{\partial \varphi_{uk}}\left[\delta_{u=v}\varphi_{uk} - \varphi_{vk}\varphi_{uk}\right]$$
$$= M_k^{-1}(1 - V\varphi_{vk}). \quad (24)$$

Since $\mathbf{G}(\boldsymbol{z})$ is block-diagonal with its $\boldsymbol{r}$-relevant block being $\mathbf{I}(\boldsymbol{r}) = M^{(L+1)}\text{diag}(1/\boldsymbol{r})$, according to (3), it is straightforward to show that

$$\Gamma_k(\boldsymbol{r}) = \sum_u \frac{\partial}{\partial r_u}\left[\mathbf{I}_{ku}^{-1}(\boldsymbol{r})\right],$$
$$= \sum_u \frac{\partial}{\partial r_u}\left[\delta_{u=k}\frac{r_u}{M^{(L+1)}}\right], \quad (25)$$
$$= 1/M^{(L+1)}.$$

## C. Proof of Lemma 3.1

Note that the counts in $x_{vj}^{(l)} \sim \text{Pois}\left(q_j^{(l)}\sum_{k=1}^{K_l}\phi_{vk}^{(l)}\theta_{kj}^{(l)}\right)$ can be augmented as

$$x_{vj}^{(l)} = \sum_{k=1}^{K_l} x_{vkj}^{(l)},$$
$$x_{vkj}^{(l)} \sim \text{Pois}(q_j^{(l)}\phi_{vk}^{(l)}\theta_{kj}^{(l)}), \quad (26)$$

which, according to Lemma 4.1 of Zhou et al. (2012), can be equivalently expressed as

$$\left(x_{vkj}^{(l)}\right)_v \sim \text{Mult}\left(m_{kj}^{(l)(l+1)}, \boldsymbol{\phi}_k^{(l)}\right),$$
$$m_{kj}^{(l)(l+1)} \sim \text{Pois}\left(q_j^{(l)}\theta_{kj}^{(l)}\right), \quad (27)$$

where $m_{kj}^{(l)(l+1)} := \sum_{v=1}^{K_{l-1}} x_{vkj}^{(l)}$. Marginalizing out $\theta_{vj}^{(l)} \sim \text{Gam}\left(\sum_{k=1}^{K_{l+1}}\phi_{vk}^{(l+1)}\theta_{kj}^{(l+1)}, 1/c_j^{(l+1)}\right)$ from (27) leads to

$$m_{vj}^{(l)(l+1)} \sim \text{NB}\left(\sum_{k=1}^{K_{l+1}}\phi_{vk}^{(l+1)}\theta_{kj}^{(l+1)}, p_j^{(l+1)}\right), \quad (28)$$

which can be augmented as

$$m_{vj}^{(l)(l+1)} \sim \text{SumLog}(x_{vj}^{(l+1)}, p_j^{(l+1)}),$$
$$x_{vj}^{(l+1)} \sim \text{Pois}\left(q_j^{(l+1)}\sum_{k=1}^{K_{l+1}}\phi_{vk}^{(l+1)}\theta_{kj}^{(l+1)}\right). \quad (29)$$

When $l = L$, we have
$$m_{kj}^{(L)(L+1)} \sim \text{NB}(r_k, p_j^{(L+1)}), \quad (30)$$

marginalizing the gamma process $G \sim \text{GaP}(G_0, 1/c_0)$ from which leads to a gamma-negative binomial process random count matrix, as expressed in the first two lines of (6).

## D. Corollary D.1

**Corollary D.1.** *The gamma-negative binomial process PFA can be equivalently expressed as*

$$\ell_{k\cdot} \sim \text{Log}\left(\frac{q_\cdot}{c_0 + q_\cdot}\right), \ K \sim \text{Pois}\left(\gamma_0 \ln \frac{c_0 + q_\cdot}{c_0}\right),$$
$$\mathcal{L} = \sum_{k=1}^{K} \ell_{k\cdot}\delta_{\boldsymbol{\phi}_k},$$
$$(\ell_{kj})_j \sim \text{Mult}\left[\ell_{k\cdot}, (q_j)_j/q_\cdot\right],$$
$$m_{kj} \sim \text{SumLog}(\ell_{kj}, p_j)$$
$$x_{vj} = \sum_{k=1}^{K} x_{vkj}, \ (x_{vkj})_v \sim \text{Mult}(m_{kj}, \boldsymbol{\phi}_k). \quad (31)$$

# E. Visualizations of the extracted topics and networks

In the following, we provide some example results, obtained using DLDA where $[K_1, K_2, K_3] = [128, 64, 32]$ and $\eta^{(l)} = 1/K_l$ for the $l$th layer, on extracting multilayer representations/topics from both the RCV1 and Wiki corpora. Clearly interpretable results, which are similar to those reported in Zhou et al. (2016a) and hence omitted here for brevity, are also extracted from the 20Newsgroups corpus.

## E.1. RCV1

Following the visualization techniques in Zhou et al. (2016a), we plot 54 example topics of layer one in Figure 4, the top 30 topics of layer two in Figure 5, and the top 30 topics of layer three in Figure 6. Figure 4 clearly shows that the topics of layer one are very specific. For example, topics 41, 71 and 62 in the first row are about "Germany," "Polish," and "France," respectively; topics 53 and 54 in the second row are about "airline" and "European union," respectively; and topics 85 and 36 in the third row are about "ship & island" and "comput & techn," respectively. By contrast, the topics of layers two and three, shown in Figures 5 and 6, respectively, are increasingly more general. Such topics can be better interpreted via the following informative tree structured visualizations. Note that a tree defined in this paper allows a child node of a layer to be connected to more than one parent node of the adjacent higher layer.

Shown in Figure 7 is a [10, 3, 1] tree rooted at node 4 of the top layer on "bonds, rates, & credit markets." Apparently, the topics become more and more specific when moving from top to bottom following the branches. For example, the root node splits into three nodes from layers three to two, which focus differently on "treasury bill," "dollar rate," and "bond, credit, & debt," respectively. When moving from layers two to one, all three topics in layer two splits into multiple ones that is clearly more specific. For example, topics 1, 17, and 87 are about "months," "loan & credit," and "bond & pay," respectively.

Shown in Figure 8 is another analogous tree rooted at node 24 of layer three. It is clear that, as the nodes of this tree, topics 55, 38, 34, and 30 of layer two are mainly about "Germany," "France," "airline," and "labor union," respectively. Moreover, these four topics at layer two are all connected to topic 8 of layer one, which is very specific on "office meetings."

To understand the relationships and distinctions between different trees, we construct subnetworks as shown in Figures 9-10. Figure 9 clearly shows that all three trees, rooted at nodes 16, 10, and 17 of layer three, respectively, are highly related to topic 3 of layer two on "low & expect". However, the two trees rooted at node 10 and 17, respectively, both have their own specificities. For example, topic 52 of layer two on "wall street," is unique to node 10 of layer three, and topic 35 of layer two on "India" is unique to node 17 of layer three. Similar phenomena can also be observed from another subnetwork on "China," shown in Figure 10, where both nodes of the top layer are connected to topic 19 of layer two on "corp & techn," topic 36 on "China," and topic 12 on "profit & expect." Though related to each other, the tree rooted at node 18 of the top layer is also strongly connected to topic 31 on "project" and topic 34 on "airline," whereas the other one focuses differently on topic 49 on "car & Korea" and topic 44 on "growth rate".

## E.2. Wiki

What follows are analogous figures illustrating some interpretable topics extracted from Wiki.

Figures 11-13 show the top example topics at layers one, two, and three, respectively. It is obvious that topics of layer one are specific, such as topic 31 on "university & research," topic 72 on "news, magazine, & times," topic 83 on "military & army," topic 74 on "police, crime & prison," topic 75 on "birds & species," topic 36 on "British & England," and so on. By contrast, when going to higher layers, topics become more general, as shown in Figures 12 and 13. To better illustrate topics of higher layers, we explicitly show their hierarchical structures via the following trees and subnetworks.

Figure 14 shows a tree rooted at node 1 of layer three on "music & song," whose topics at layer two are about "song & band" and "music, piano, & theatre," respectively. Figure 15 demonstrates another tree consisting of topic 9 of layer two on "London & British," topic 50 on "church & Catholic," and topic 25 on "king & prince," which is mainly about "United Kingdom." Given in Figure 16 is another tree on "art & museum," where the left side is about "art" while the right is on "history & building." These trees are all clearly interpretable.

In the subnetwork shown in Figure 17, all three trees are related to topic 9 of layer two on "London, British, & Sir." But they focus differently on topic 16 of layer two on "Irish Americans," topic 24 on "life, birth, education, career, family, & death," and topic 25 on "king & prince," respectively. Similar phenomena can also be observed in Figure 18. Both trees are related to topic 52 of layer two on "ship" and topic 49 on "air," but the left one is about various means of transportation and communication while the right one is about various components of "war." Figure 19 shows another subnetwork on "team & race," where three trees, all include topic 6 of layer two, focus differently on "goals, clubs, & league," "world cup," and "rank, first, second, & third," respectively.

**Deep Latent Dirichlet Allocation with TLASGR MCMC**

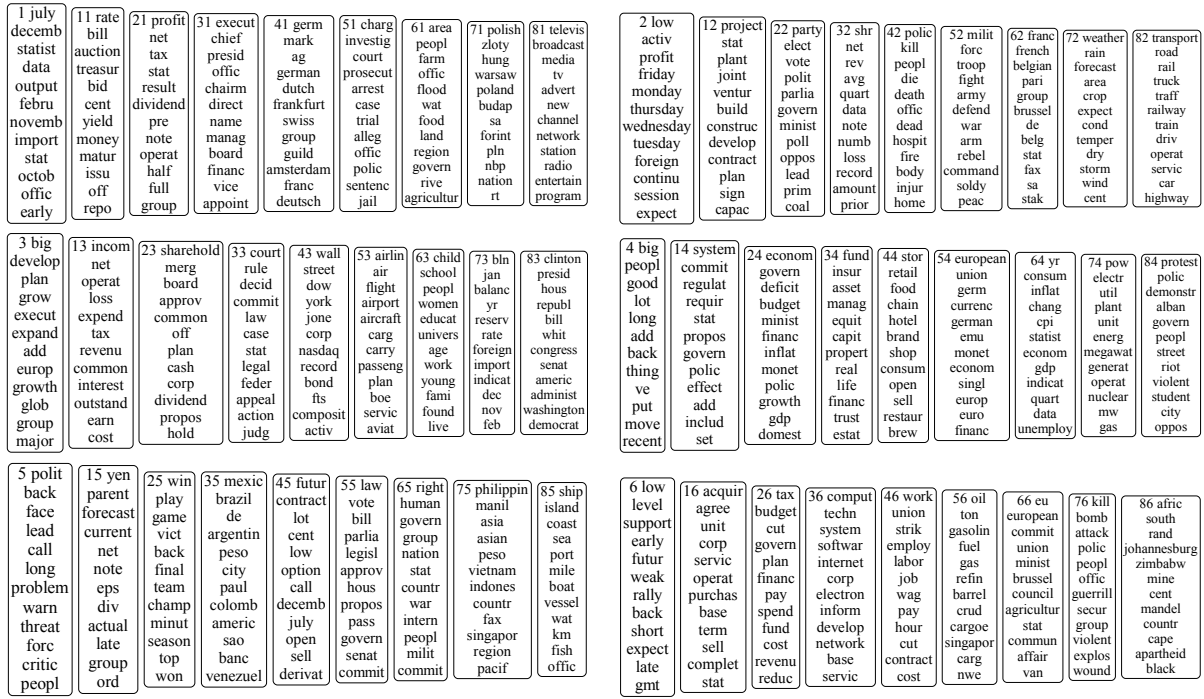

Figure 4. Example topics of layer one of DLDA trained with TLASGR MCMC on RCV1.

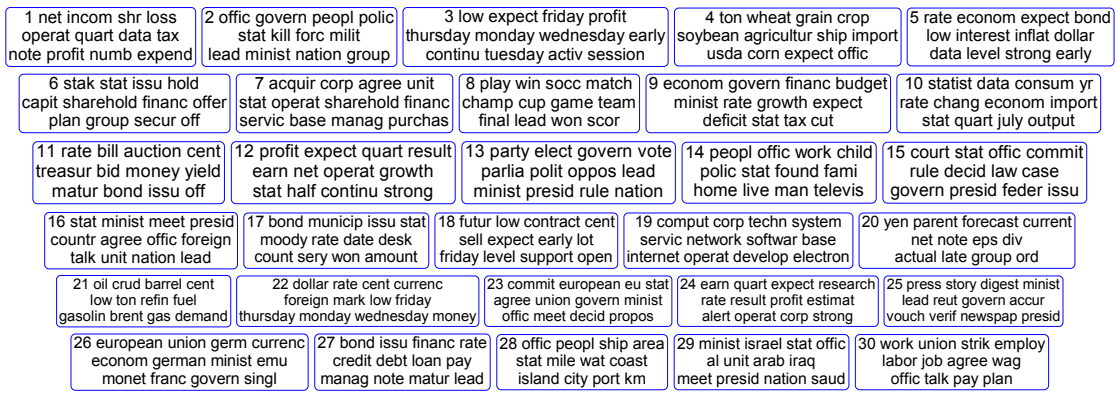

Figure 5. The top 30 topics of layer two of DLDA trained with TLASGR MCMC on RCV1.

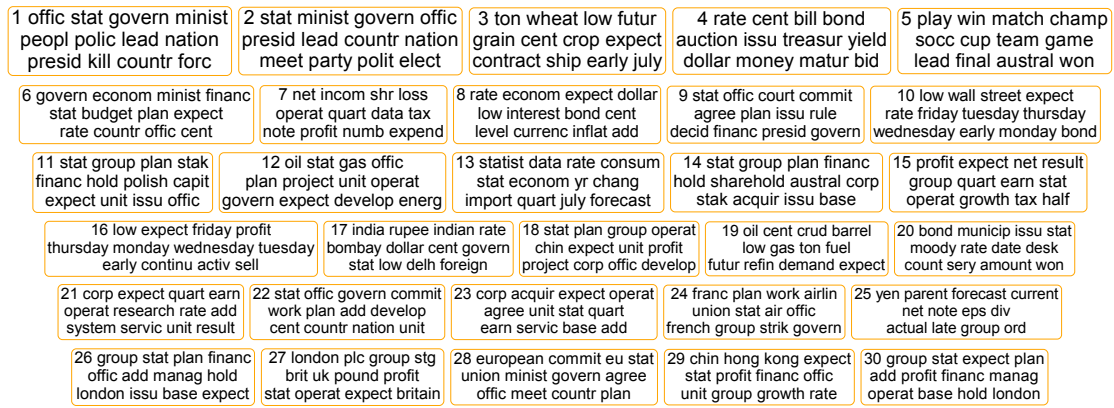

Figure 6. The top 30 topics of layer three of DLDA trained with TLASGR MCMC on RCV1.

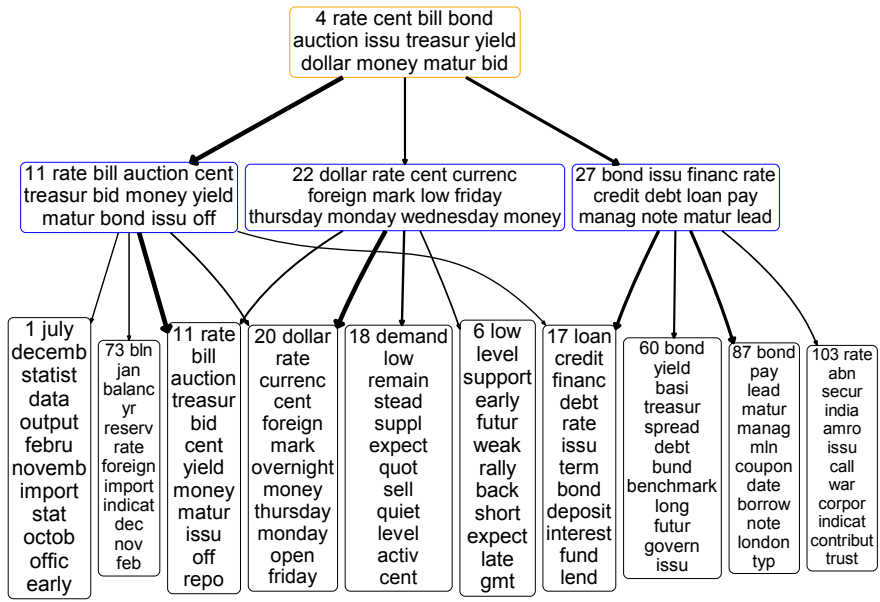

*Figure 7.* A $[10, 3, 1]$ tree on "bonds, rates, & credit markets" that includes all the lower-layer nodes (directly or indirectly) linked with non-negligible weights to node 4 of the top layer, taken from the full $[128, 64, 32]$ DLDA network trained with TLASGR MCMC on the 794,414 training documents of the RCV1 corpus, with $\eta^{(l)} = 1/K_l$ for the $l$th layer. A line from node $k$ at layer $l$ to node $k'$ at layer $l - 1$ indicates that $\mathbf{\Phi}^{(l)}(k', k) > 5/K_{l-1}$, with the width of the line proportional to $\sqrt{\mathbf{\Phi}^{(l)}(k', k)}$. For each node, the rank (in terms of popularity) at the corresponding layer and the top 12 words of the corresponding topic are displayed inside the text box, where the text font size monotonically decreases as the popularity of the node decreases, and the outside border of the text box is colored as orange, blue, or black if the node is at layer three, two, or one, respectively.

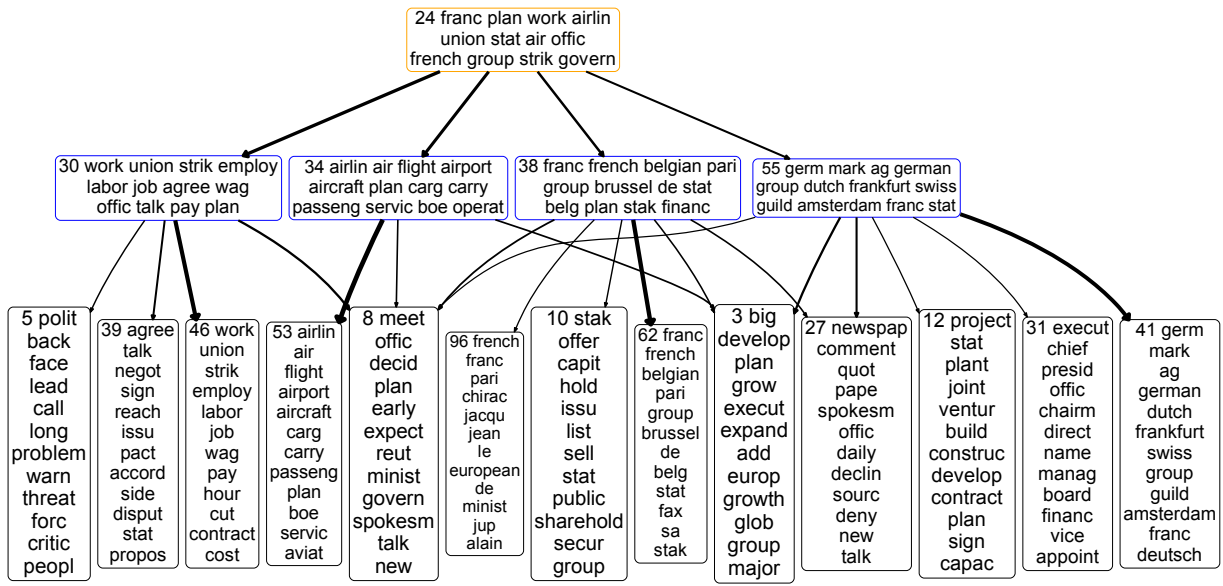

*Figure 8.* Analogous plots to Figure 7 for a tree rooted at node 24 on "France, Germany, & airline" from RCV1.

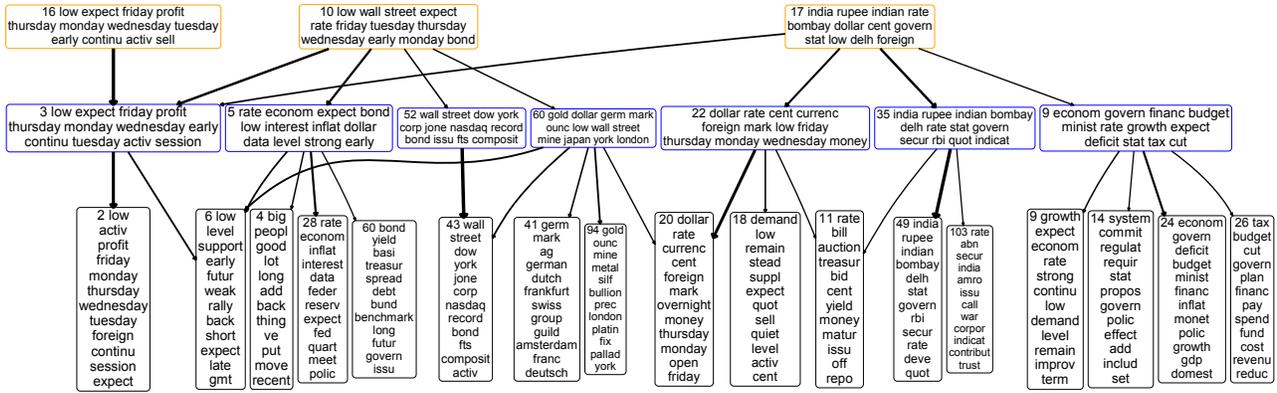

*Figure 9.* Analogous plots to Figure 7 for a subnetwork related to "low & expect" from RCV1, consisting of three trees rooted at nodes 16, 10 and 17, respectively, of layer three.

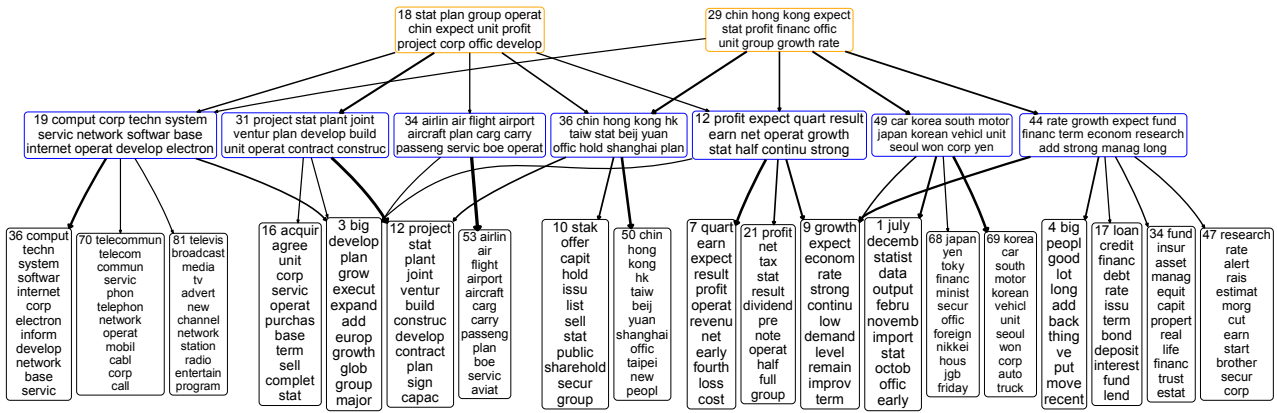

*Figure 10.* Analogous plots to Figure 7 for a subnetwork on "China" from RCV1, consisting of two trees rooted at nodes 18 and 29, respectively, of layer three.

# Deep Latent Dirichlet Allocation with TLASGR MCMC

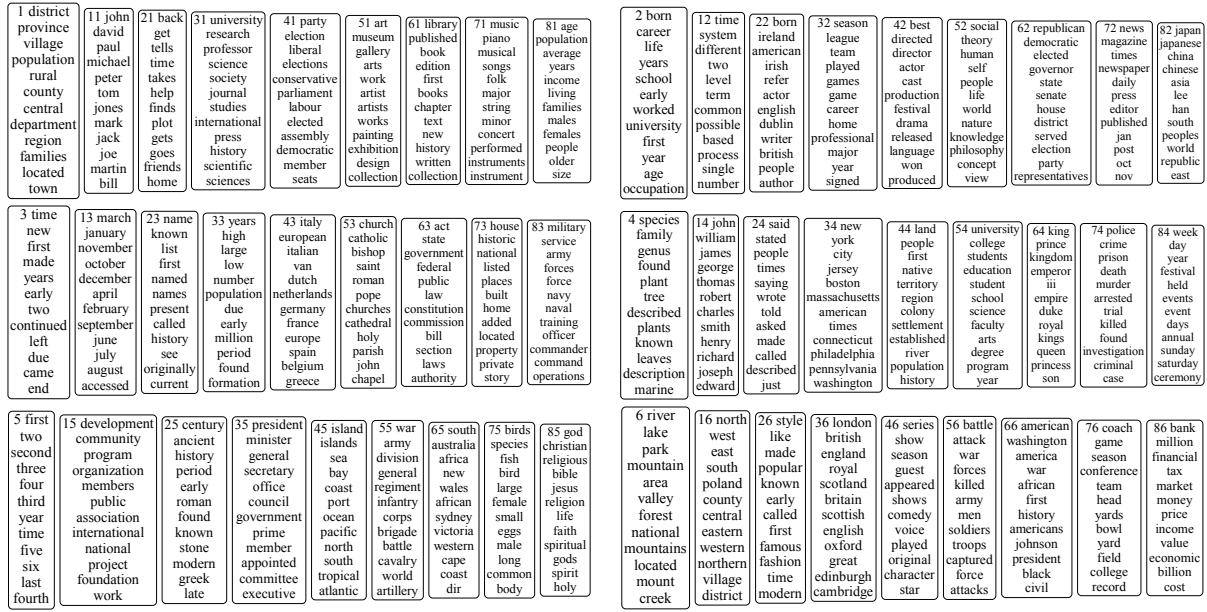

Figure 11. Example topics of layer one of DLDA trained with TLASGR MCMC on Wiki.

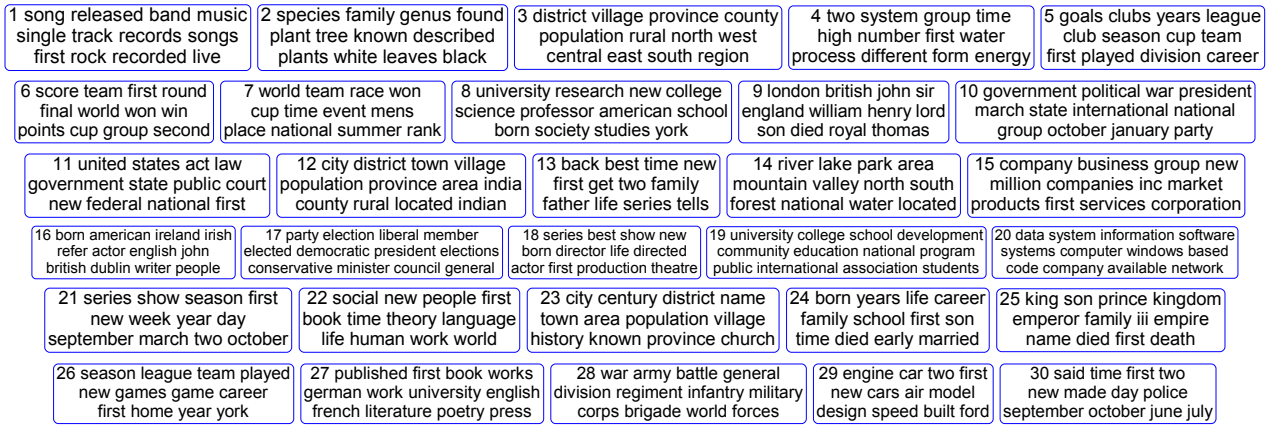

Figure 12. The top 30 topics of layer two of DLDA trained with TLASGR MCMC on Wiki.

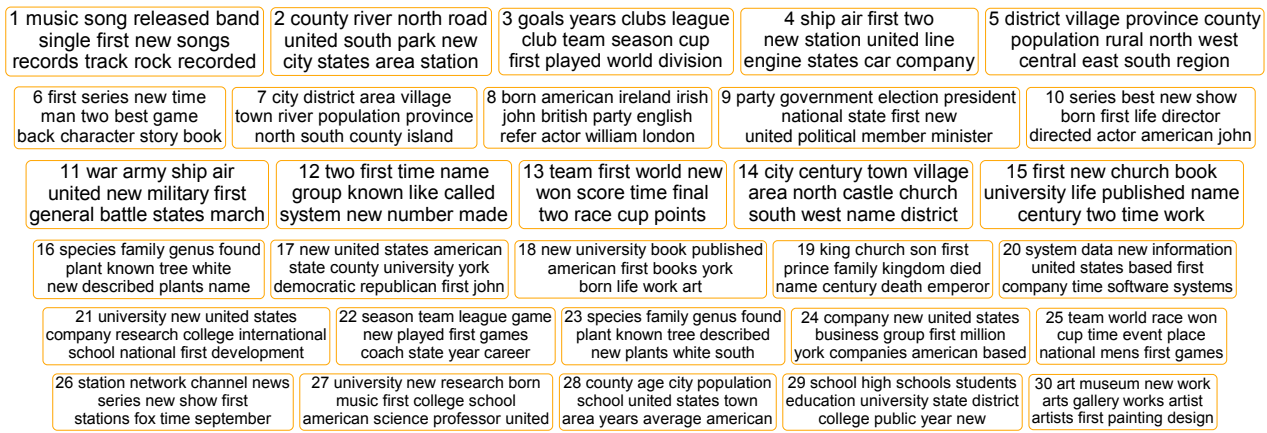

Figure 13. The top 30 topics of layer three of DLDA trained with TLASGR MCMC on Wiki.

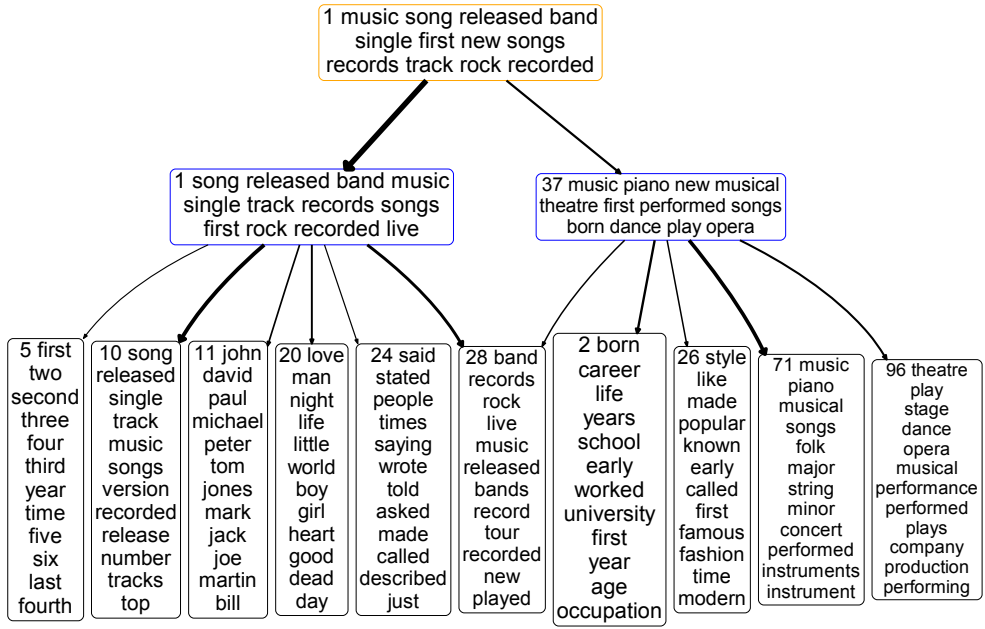

*Figure 14.* Analogous plots to Figure 7 for a tree rooted at node 1 on "music & song" from Wiki.

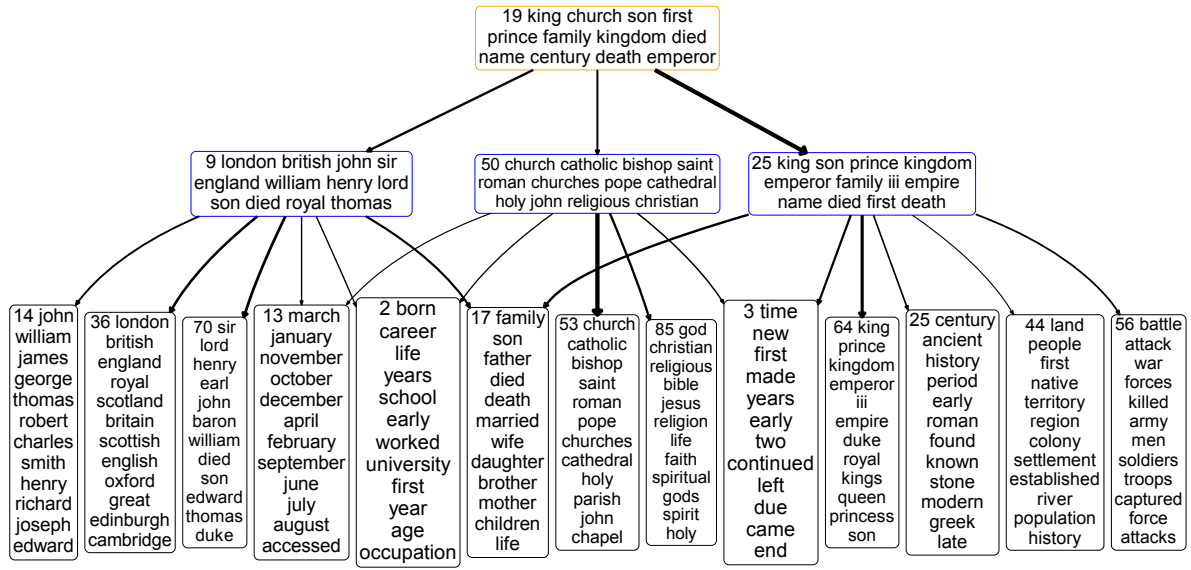

*Figure 15.* Analogous plots to Figure 7 for a tree rooted at node 19 on "United Kingdom" from Wiki.

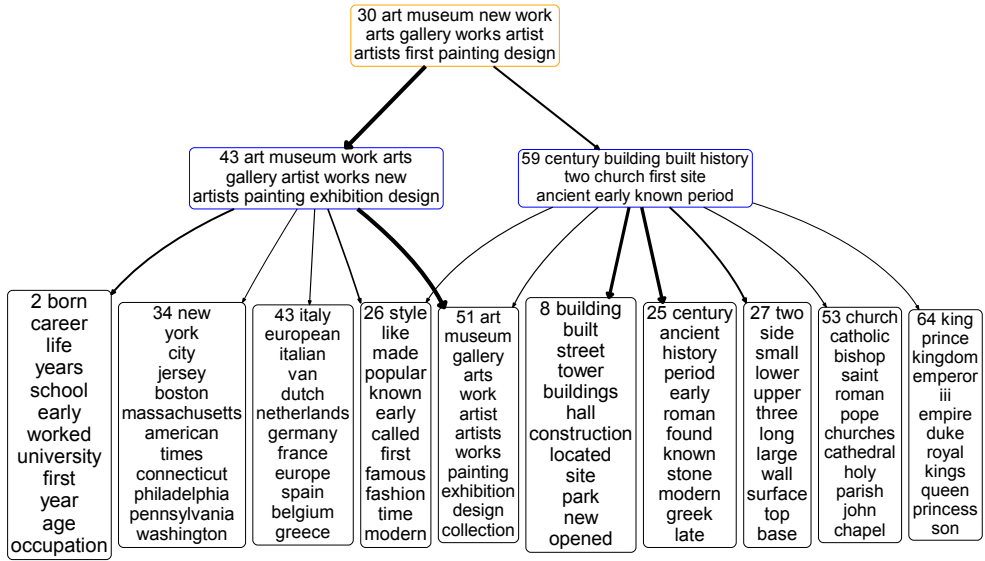

*Figure 16.* Analogous plots to Figure 7 for a tree rooted at node 30 on "art & museum" from Wiki.

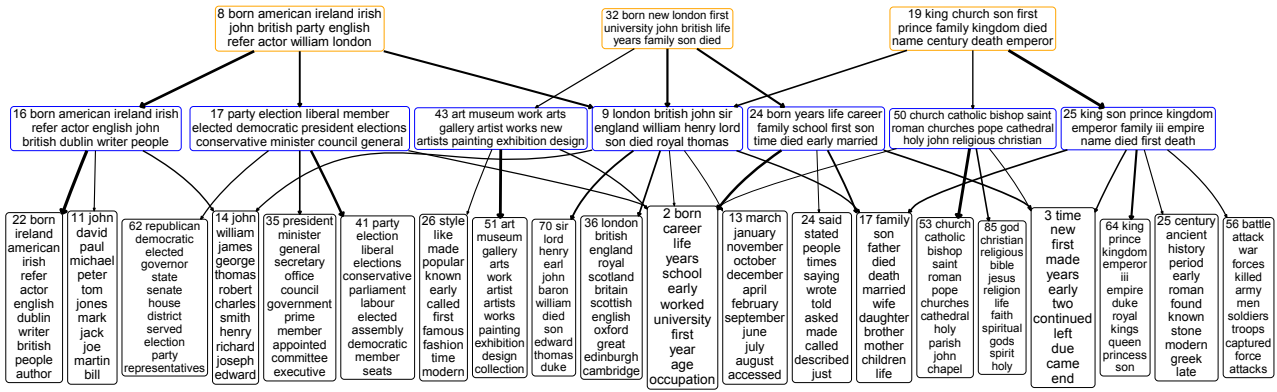

*Figure 17.* Analogous plots to Figure 7 for a subnetwork on "British" from Wiki, consisting of three trees rooted at nodes 8, 32, and 19, respectively, of layer three.

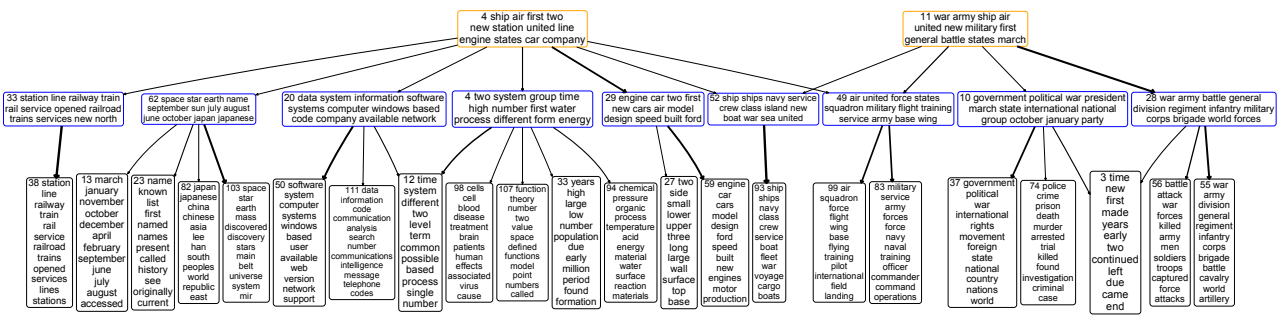

*Figure 18.* Analogous plots to Figure 7 for a subnetwork on "ship & air" from Wiki, consisting of two trees rooted at nodes 4 and 11, respectively, of layer three.

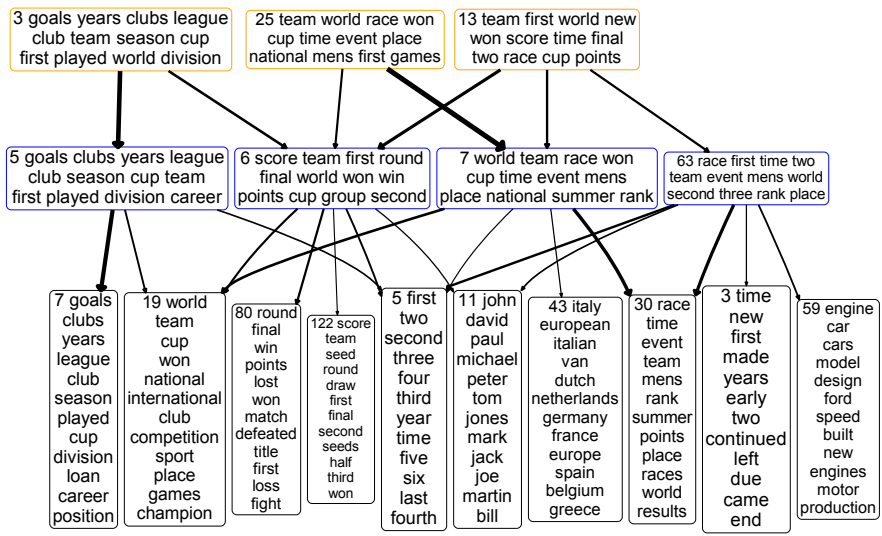

*Figure 19.* Analogous plots to Figure 7 for a subnetwork on "team & race" from Wiki, consisting of three trees rooted at nodes 3, 25, and 13, respectively, of layer three.